\documentclass[11pt,a4paper]{article}
\usepackage[hyperref]{acl2021}
\aclfinalcopy 
\usepackage{times}
\usepackage{nert}
\usepackage{latexsym}
\usepackage[T1]{fontenc}
\usepackage{bbm}
\usepackage{amsmath}
\usepackage{amsfonts}
\usepackage{xspace}
\usepackage{subcaption}

\usepackage{mathtools}

\usepackage{soul}
\expandafter\def\expandafter\UrlBreaks\expandafter{\UrlBreaks
  \do\a\do\b\do\c\do\d\do\e\do\f\do\g\do\h\do\i\do\j%
  \do\k\do\l\do\m\do\n\do\o\do\p\do\q\do\r\do\s\do\t%
  \do\u\do\v\do\w\do\x\do\y\do\z\do\A\do\B\do\C\do\D%
  \do\E\do\F\do\G\do\H\do\I\do\J\do\K\do\L\do\M\do\N%
  \do\O\do\P\do\Q\do\R\do\S\do\T\do\U\do\V\do\W\do\X%
  \do\Y\do\Z}
\usepackage{caption}
\usepackage{newfloat}
\usepackage{tikz}
\definecolor{LimeGreen}{RGB}{179,253,148}
\definecolor{OliveGreen}{rgb}{0,0.6,0}
\DeclareRobustCommand{\hlpink}[1]{{\sethlcolor{pink}\hl{#1}}}
\DeclareRobustCommand{\hlgreen}[1]{{\sethlcolor{green}\hl{#1}}}

\usepackage{adjustbox}
\usepackage{array}
\usepackage{color, colortbl}
\newcolumntype{R}[2]{%
    >{\adjustbox{angle=#1,lap=\width-(#2)}\bgroup}%
    l%
    <{\egroup}%
}
\def\basiceval#1{\the\numexpr#1\relax}
\def\cca#1{\cellcolor{blue!#10}\ifnum #1>4\color{white!#10}\else\color{black!100}\fi{.#1}}
\def\ccb#1{\cellcolor{red!#10}\ifnum #1>4\color{white!#10}\else\color{black!100}\fi{-.#1}}
\def\ccaten#1{\cellcolor{blue!#1}\ifnum #1>40\color{white!100}\else\color{black!100}\fi{
\ifnum #1>10 .#1 \else .0#1 \fi
}}
\def\ccatentwo#1#2{\cellcolor{blue!\basiceval{#2*10}}\ifnum #2>4\color{white!#1}\else\color{black!100}\fi{#1.#2}}
\def\ccbten#1{\cellcolor{red!#1}\ifnum #1>40\color{white!#1}\else\color{black!100}\fi{
\ifnum #1>10 -.#1 \else -.0#1 \fi
}}
\def\ccbtentwo#1#2{\cellcolor{red!\basiceval{#2*5}}\ifnum #2>4\color{white!#1}\else\color{black!100}\fi{-#1.#2}}
\def\ccd#1{\cellcolor{blue!\basiceval{#1 * 2}}\ifnum #1>20\color{white!#1}\else\color{black!100}\fi{
\ifnum #1>10 .#1 \else .0#1 \fi
}}
\def\ccy#1{\cellcolor{yellow!\basiceval{#1 * 3}}\ifnum #1>50\color{white!#1}\else\color{black!100}\fi{
\ifnum #1>9 .#1 \else .0#1 \fi
}}

\definecolor{darkgreen}{HTML}{38761d}
\definecolor{darkyellow}{HTML}{ffa500}
\definecolor{brightlavender}{rgb}{0.75, 0.58, 0.89}
\DeclareRobustCommand{\hlcyan}[1]{{\sethlcolor{cyan}\hl{#1}}}
\DeclareRobustCommand{\hlpink}[1]{{\sethlcolor{pink}\hl{#1}}}
\DeclareRobustCommand{\hlyellow}[1]{{\sethlcolor{yellow}\hl{#1}}}
\DeclareRobustCommand{\hlpurple}[1]{{\sethlcolor{brightlavender}\hl{#1}}}
\DeclareRobustCommand{\hlgreen}[1]{{\sethlcolor{green}\hl{#1}}}

\usepackage{microtype}

\title{\textit{StateCensusLaws.org}: A Web Application for Consuming and Annotating Legal Discourse Learning}

\author{Alexander Spangher and Jonathan May \\ 
Information Sciences Institute \\
  University of Southern California \\ 
  \texttt{spangher@usc.edu}, \texttt{jonmay@isi.edu} \\}

\begin{document}
\maketitle
\begin{abstract}
In this work, we create a web application to highlight the output of NLP models trained to parse and label discourse segments in law text. Our system is built primarily with journalists and legal interpreters in mind, and we focus on state-level law that uses U.S. Census population numbers to allocate resources and organize government. 
    
Our system exposes a corpus we collect of $6,000$ state-level laws that pertain to the U.S. census, and highlights our discourse parsing model's output. We also build in a novel, flexible annotation framework that can handle span-tagging and relation tagging on an arbitrary input text document. This framework can allow journalists and interested practitioners to correct tags, and tag more documents.
\end{abstract}

\section{Introduction}

Since at least 1958, AI practitioners have explored how to analyze legal documents -- i.e. laws, court opinions and regulations -- to yield greater insight into legal decision-making \cite{mehl1958automation}. A number of systems seek to help citizens understand law \cite{dale2019law}, by answering questions\footnote{\url{https://www.chatbotsecommerce.com/nrf-launches-parker-first-australian-privacy-law-chatbot/}}, generating documents\footnote{\url{https://legal.thomsonreuters.com.au/products/contract-express/}, \url{https://turbotax.intuit.com/}}, or helping users file motions \cite{gibbs2016}.

However, researchers trying to compare and contrast large bodies of law, such as journalists or academics, have virtually no open-source resources: an informal survey of legal journalists we conducted before starting this project\footnote{Many were former colleagues at the \textit{New York Times} or current colleagues at the USC Annenberg School for Communication and Journalism.} yielded several insights: (1) there are no open accessible, free, online sources to perform full-text searches on all state law (2) it is hard to track entities across laws (3) it is hard to know when a law applies. Our web application aims to address all three of these short-comings.

\begin{figure}[t]
    \centering
    \fbox{
        \parbox{.45\textwidth}{
            ...in \hlpink{counties} \hlyellow{having a metropolitan form of government} and in \hlpink{counties} \hlyellow{having a population of not less than three hundred thirty-five thousand (335,000) nor more than three hundred thirty-six thousand (336,000), according to the 1990 federal census or any subsequent federal census}, the \hlcyan{magistrate or magistrates} \hlpurple{shall be selected and appointed by and serve at the pleasure of} \hlgreen{the trial court judge}...
        }
    }
    \caption{Paragraph from a sample law, Tennessee § 36-5-402, referencing a bureaucratic process impacted by population counts determined by the upcoming federal census. The colored blocks represent the following concepts, which we will define in the text: \hlpink{probe}, \hlyellow{test}, \hlgreen{subject}, \hlpurple{consequence}, \hlcyan{object}. Our web application aggregates these span tags across state-level laws, helping civil citizens and journalists more easily discern the impacts of a U.S. Census undercount.}
    \label{fig:example_law}
\end{figure}

In this work, we present a web application\footnote{\url{http://www.statecensuslaws.org/}}, an annotation framework, and a set of web-scrapers designed to meet these challenges. This system is the fruit of theoretical work, publication-upcoming, in which we propose a discourse schema for analyZing law (illustrated in Figure \ref{fig:example_law}), an annotated dataset, and a set of models. At the core, our framework seeks to answer the following key questions: (1) When does this law apply? (2) Who gains what powers? (3) Who gains what restrictions? This paper makes three key contributions:
\begin{enumerate}
    \item \textbf{Searching and consuming model output}: We present a web-app to expose and help users navigate through a discourse framework for law that we develop, annotate and train models on.
    \item \textbf{Collecting and Updating Annotations}: We develop a light-weight and modular Javascript-based annotation interface for span and relation-labeling to integrate into webpages, standalone apps, or Amazon Mechanical Turk (AMT) tasks. 
    \item \textbf{Web Scraping Public Domain U.S. State Law}: We release a set of 25 robust Dockerized web-scrapers to collect U.S. state law wholesale or by keyword. These scrapers are designed to overcome uncivil attempts to block law-scraping that hinders research. We collect and release a database of more than $6,000$ state-level laws using these scrapers.
\end{enumerate}

Our particular subject area in this work -- state-level laws pertaining to the 2020 U.S. Census -- is relevant for several reasons: (1) The 2020 U.S. Census has faced massive challenges and certain populations might be undercounted \cite{naylor2020counting, mervis2019census, berry2020civil}. (2) Very little research exists exploring the effects of an undercount on state-level processes.\footnote{Research exists illuminating the effects of populations undercounts on Federal budgeting and Congressional representation \cite{reamer2018counting, berry2020civil}, but because of the challenges discussed, state-level law is harder to probe.} (3) Journalists, our primary users, can provide useful feedback for ongoing work on discourse-schema development.

We outline our discourse schema and modeling in Section~\ref{sct:background}. We next discuss our dataset collection process, including the web-scrapers we release for gathering public-domain U.S. state law text (Section \ref{sct:fulldatasetcollection}). In Section \ref{sct:annotationinterface} we describe our lightweight and modular span and relation annotation interface which we used to collect data. Next, in Section \ref{sct:webinterface}, we describe our web-app, where we surface our model's output to journalists and engage volunteers to improve our annotations. Finally, we discuss an ongoing use-case to illustrate how one might use our app in Section \ref{sct:usecase}.


\section{Background}
\label{sct:background}

We briefly introduce the key components of our discourse schema, the dataset we collect, and the BERT-CRF-based modeling we perform. 

\subsection{Schema}
\label{subsct:schema}
The five principal discourse elements that we identify in our schema are: \hlgreen{SUBJECT}, \hlpurple{CONSEQUENCE}, \hlcyan{OBJECT}, \hlpink{PROBE} and \hlyellow{TEST}. PROBES, SUBJECTS and OBJECTS are entities while TESTS and CONSEQUENCES are verb phrases. We describe each in turn.

The first three elements in our schema, SUBJECT, CONSEQUENCE, and OBJECT, uncover how law dictates first-degree interactions between entities\footnote{
Inspired by seminal work done by \newcite{gardner1984artificial}.
}.
As such, a SUBJECT is an entity directly gaining a power or a restriction under a law. The CONSEQUENCE is the specific power or restriction placed on the SUBJECT. The OBJECT is the entity being affected by the SUBJECT's gain in powers or restrictions.\footnote{SUBJECT, CONSEQUENCE and OBJECT example: \textit{\hlgreen{The trial court judge} \hlpurple{shall adjudicate property disputes between} \hlcyan{claimants}.}} 
OBJECTS are not always present,\footnote{Example of a SUBJECT-CONSEQUENCE relation without an OBJECT: \textit{\hlgreen{The trial court judge} \hlpurple{shall begin session at or before 9am.}}} 
and SUBJECTS might be expressed passively.\footnote{Passive SUBJECT (and passive OBJECT) example: \textit{\hlpurple{Taxes shall be collected at the beginning of every month.}} The SUBJECT and OBJECT, \hlgreen{``Tax-collector''} and \hlcyan{``Tax-payer''}, are not actively expressed. (SUBJECTS and OBJECTS are nearly always individuals or organizations).}

The last two elements in our schema, TEST, and PROBE, indicate when laws apply. A TEST is an explicit condition applied to an entity to determine when a SUBJECT-CONSEQUENCE-OBJECT relation holds. A PROBE is an entity that is tested but not part of a legally-mediated relationship.\footnote{PROBE and TEST example: \textit{In \hlpink{counties} \hlyellow{with a population above 10,000}, the trial court judge shall adjudicate property disputes between claimants.}} A TEST need not only apply to a PROBE; it can apply to a SUBJECT, OBJECT, or even a CONSEQUENCE.

\subsection{Annotated Dataset and Modeling}

Our dataset collection is described in Section \ref{sct:fulldatasetcollection}. From our full dataset, we sample a set of law-paragraphs to annotate.
 We build an annotation framework, described in Section \ref{sct:annotationinterface}, and enlist two expert annotators. As of this writing, we have 573 law paragraphs annotated according to the span-level schema described in Section~\ref{subsct:schema}.

We test a class of models aimed at performing span-tagging, and use our top-performing model, BERT-CRF (based on \newcite{li2019discourse, spangher2021multitask}) in the current production pipeline. 
Our model accuracy improves with more data, and 
a primary purpose of our web-app is to collect more annotations from users (Section \ref{sct:annotationinterface}). We provide our predicted outputs in the database as well.

\section{Dataset Construction}
\label{sct:fulldatasetcollection}

Our collected dataset comprises the more than 100,000 active state-level laws in the United States, including roughly 6,000 laws that reference Census population counts from a public-domain law website, Justia,\footnote{\url{https://www.justia.com/}} and directly from state websites.\footnote{Some of the laws provided by Justia, such as those for Colorado, contain data in unhelpful forms, such as PDF files (see \url{https://law.justia.com/codes/colorado/2019/}), so we extract directly in these cases.} In total, we build parsers for 24 official state websites and one for Justia.

State law is \textit{always} public domain;\footnote{\url{https://fairuse.stanford.edu/overview/public-domain/welcome/}} yet in practice, it is often inaccessible for bulk downloads and web scraping. For instance, many websites license LexisNexis, a for-profit company, as the official provider for their state codes\footnote{Ex. Colorado, Georgia and Tennessee: \url{http://www.lexisnexis.com/hottopics/colorado}, \url{http://www.lexisnexis.com/hottopics/gacode}, \url{http://www.lexisnexis.com/hottopics/tncode}}. Although these websites are publicly and freely accessible (as apart from other LexisNexis-hosted resources, which require memberships to access), they employ a range of mechanisms (e.g. timeouts, dynamically-generated URLs, cookie-based access) that make them difficult to scrape.\footnote{The practical effect of mechanisms to block bulk downloads is the hindrance of law corpora collection for journalistic or academic study, where computation might be applied either to search, parse or compare the laws.}

To circumvent these blocking mechanisms, our scrapers are robust and, in many cases, mimic human web-browsing behavior in order to provide full access to public-domain law. We develop a generalized scraper for LexisNexis Public Access websites that uses \texttt{scrapy}\footnote{\texttt{Scrapy}, \url{https://scrapy.org/}, is a popular python library for scraping that handles timeouts, retries and storage.} and \texttt{selenium-webdriver}\footnote{\texttt{selenium-webdriver}, \url{https://www.selenium.dev/} is a programmatic web-browser that is commonly used to navigate websites by mimicking human behavior and to render dynamic webpages for retrieval.} In order to scrape Justia, we launch three Google Compute Engine (GCE) instances for a total of 60 compute hours. We will open-source our scraping routines as well as the Docker images we created to perform these scrapes. Our scrapers are flexible and can download either full codes or all codes related to a query. Our Docker containers contain the necessary python libraries (e.g. \texttt{scrapy}) and binary libraries \texttt{selenium-webdriver} and \texttt{gsutil} for scraping and Google Cloud Platform (GCP) storage.  These routines constitute a considerable resource for academic inquiries into state-level law.

\begin{figure*}[t]
    \centering
    \includegraphics[trim={0cm 0cm 0cm 1.9cm},clip,width=1\linewidth]{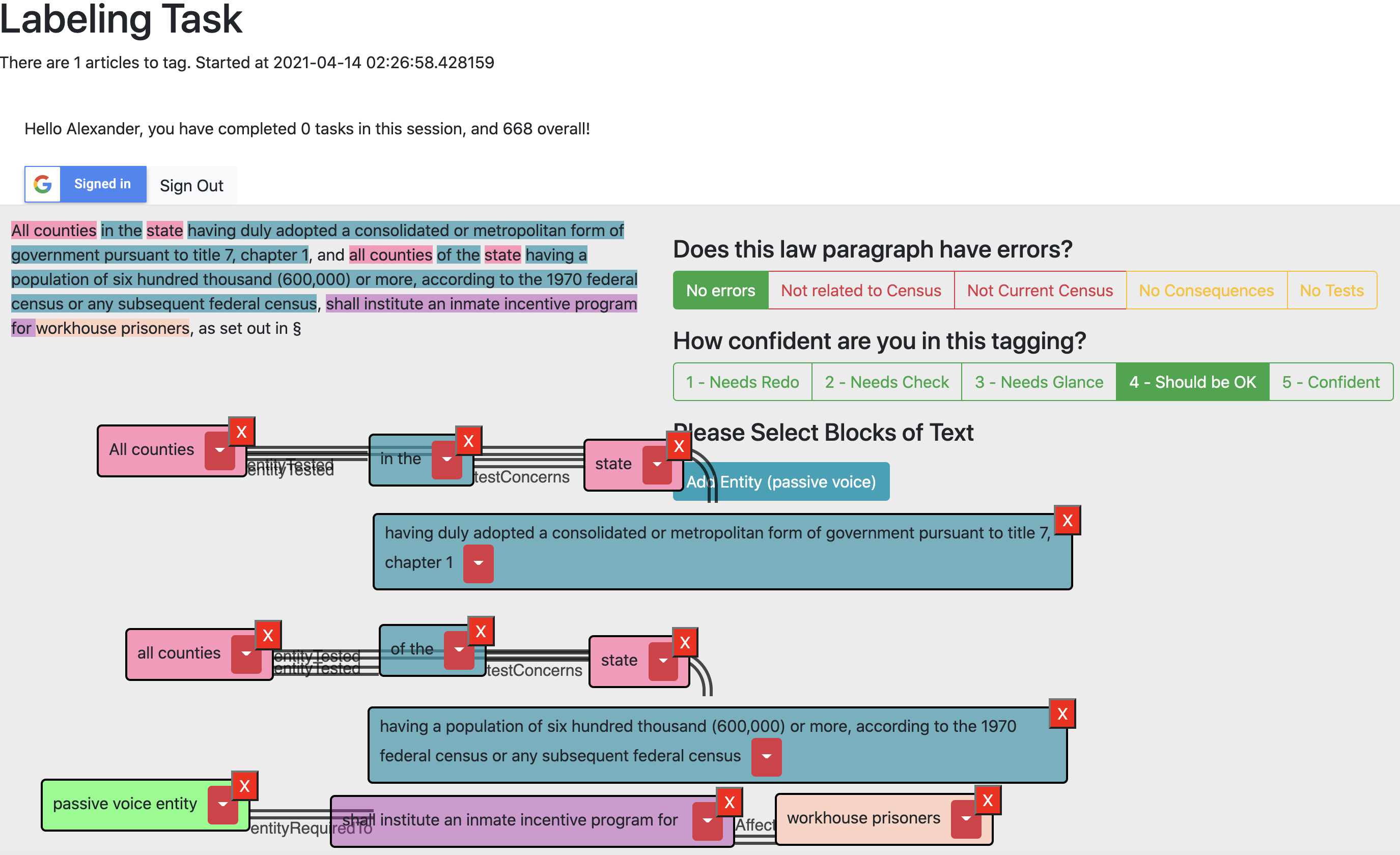}
    \caption{Our lightweight annotation interface allows users to identify spans by highlighting text, assigning labels to spans using the dropdown, and assigning relations by right-clicking to draw lines. We track user-identity using Google sign-in. (While our interface \textit{can} collect relational labels, we do not study them in this work.)}
    \label{fig:annotationinterface}
\end{figure*}

\begin{figure*}[t]
    \centering
    \includegraphics[width=1\linewidth]{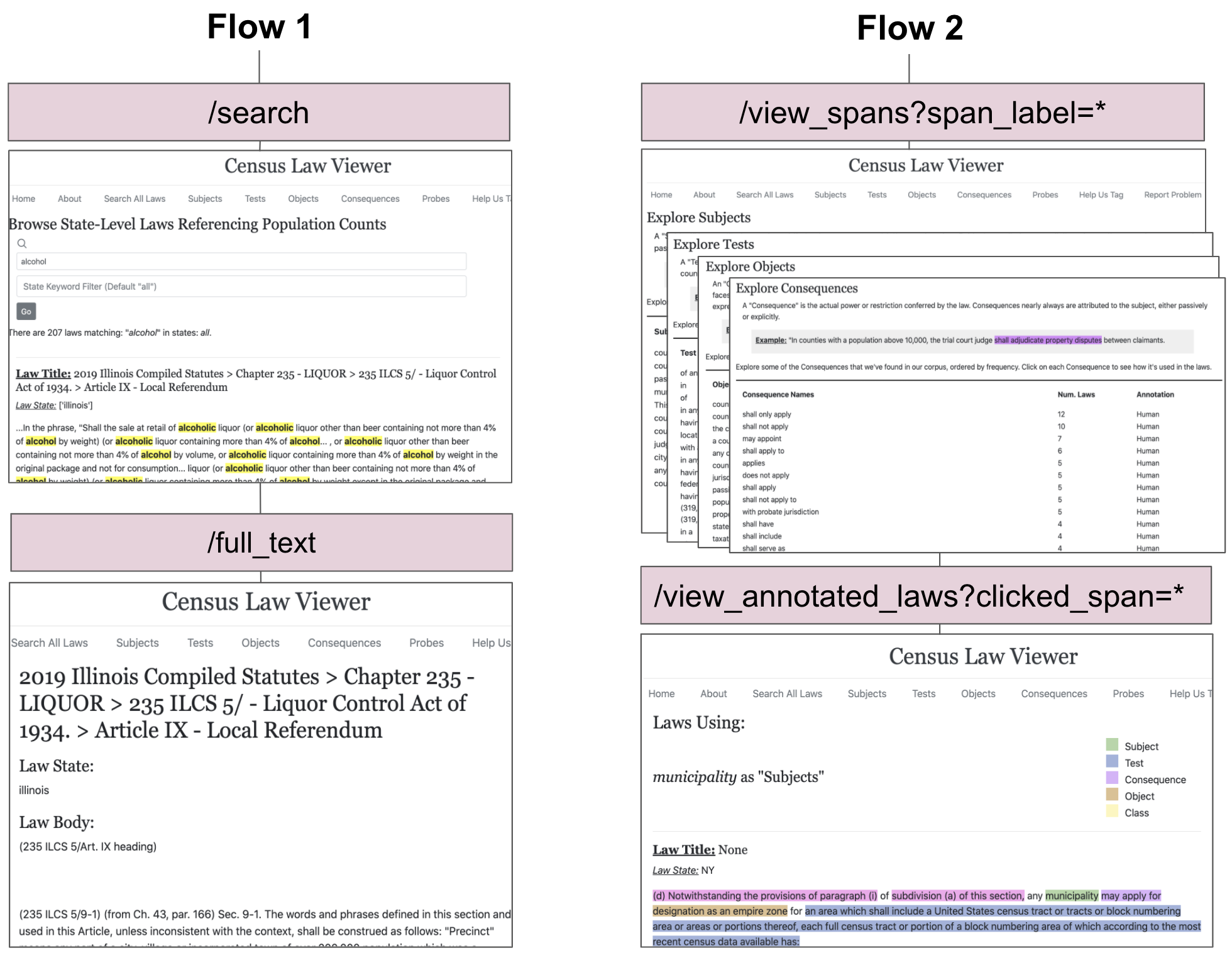}
    \caption{A flow-based sitemap for our website, \texttt{statecensuslaws.org}, with some details about the back-end and database setup. The left-column shows \textbf{Flow 1}, where a user can search and view full-text results. The right-column shows \textbf{Flow 2}, where a user can view top law-discourse spans and see all laws these spans are used in. Each flow leads to the annotation framework.}
    \label{fig:sitemap}
\end{figure*}

\section{Annotation Interface}
\label{sct:annotationinterface}

We wanted a lightweight, modular, Javascript-based annotation framework that could handle span annotation and relation assignment. We wanted it to be easily adaptable for multiple use-cases: (1) to serve as a standalone web-app to be distributed to undergraduate helpers and volunteers, (2) to be integrated into a larger web-app to help site visitors annotate new laws and correct problematic annotations they see, and (3) to be compiled to Amazon Mechanical Turk (AMT) HTMLQuestions\footnote{An HTMLQuestion is an HTML page that is submitted to AMT, \url{https://docs.aws.amazon.com/AWSMechTurk/latest/AWSMturkAPI/ApiReference_HTMLQuestionArticle.html}. AMT handles the traffic split and data collection.}.

Although many web-based, NLP-focused annotation tools exist (78, by \newcite{neves2021extensive}'s count), we surveyed options, including brat \cite{stenetorp2012brat}, GATE \cite{bontcheva2013gate}, YEDDA \cite{yang2017yedda} and WebAnnon \cite{yimam2013webanno}, and found that none of them met our requirements. Although all these options were web-based, none were flexible enough to be integrated easily into larger websites. Further, none were able to automatically generate AMT tasks.

Not finding an appropriate solution, we designed a simple and modularized annotation framework in 600 lines of JQuery, Javascript and HTML, with a Datastore backend\footnote{Google Datastore is a NoSQL, scalable JSON store, which is suitable for our usecase. \url{https://cloud.google.com/datastore}}. Our annotation framework supports span annotation and relation tagging.

\begin{figure*}[t]
    \centering
    \includegraphics[trim={0cm 0cm 0cm 0cm},clip,width=1\linewidth]{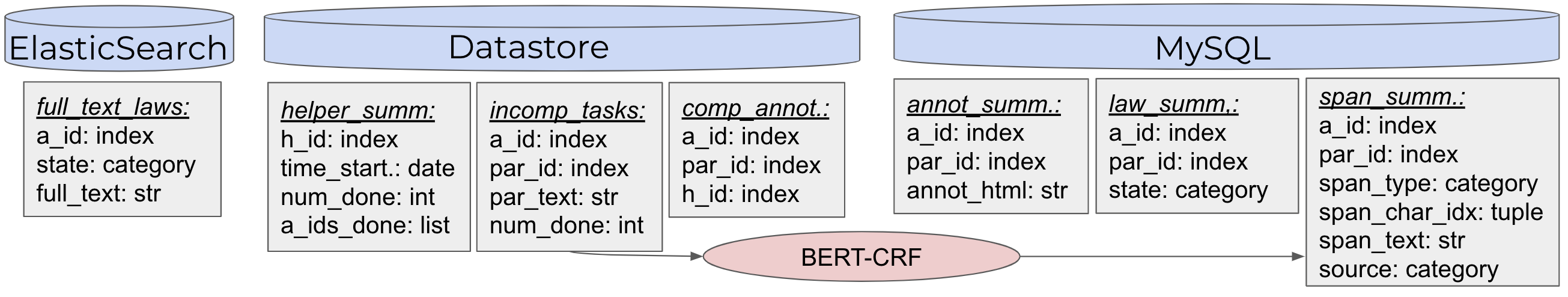}
    \caption{Our database setup. \texttt{ElasticSearch} contains the full text of laws for \textbf{Flow 1}. MySQL contains tables for summary statistics, rendering annotation HTML, and collecting annotation-data/model-output (BERT-CRF feeds into MySQL \texttt{span\_summ.}). Datastore handles task assignment for volunteers.}
    \label{fig:databaselayout}
\end{figure*}

The annotation interface itself, shown in Figure \ref{fig:annotationinterface}, is powered by a stateful page object, called \texttt{PageHandler}, that is instantiated with several parameters (\texttt{page\_height}, \texttt{buttons}, \texttt{relations}) and handles all of the page interactions. The \texttt{PageHandler} is placed directly in the HTML page containing the text to be annotated, so any service that can render text can automatically become an annotation service. In our case, we built Jinja templates to render our HTML, since our server is coded in Python-Flask. We additionally provide a helper function that, with input data, can compile our Jinja templates as static, fully-functional AMT HTMLQuestions.

We use a Datastore backend to track progress towards annotation tasks, as shown in Figure \ref{fig:databaselayout}. We code data entries (the equivalent of MySQL tables) to track helper-statistics, \texttt{helper\_summ}, how many tasks are left to assign, \texttt{incomp\_tasks}, and how many annotations have been completed, \texttt{comp\_annot}. We track these statistics to ensure that we can obtain multiple annotations for each task, and that no helper sees the same task more than once. We perform one GET request at the beginning of each user session to collect user-stats and then use client-side cookies throughout the session to minimize the number of requests we send to the back-end. We use a NoSQL database because they are low-latency and designed for streaming, and Datastore because our web-app is hosted on Google App Engine. We include our Datastore management back-end as part of the annotation package. To use our tool with other NoSQL providers,\footnote{e.g. Amazon DynamoDB -- \url{https://aws.amazon.com/dynamodb/}} a port is necessary. 

The annotation interface is designed for modularity first-and-foremost. 
We release the annotation code as part of this framework. In the future, the annotation component of this project will be abstracted and distributed in its own stand-alone Javascript package for the benefit of researchers needing a similar application.

\section{Web Interface}
\label{sct:webinterface}

We design a web interface that serves three principle use cases: (1) enabling full-text search on our database, (2) exposing users to our discourse schema by extracting spans across laws and (3) allowing users to both correct/update and provide new annotations. Our website has two principlal flows: \textbf{Flow 1}: searching laws by keyword, and \textbf{Flow 2}: grouping laws by span, shown in Figure \ref{fig:sitemap}.

In \textbf{Flow 1}, users can use a query box to perform full-text and faceted search on laws and then click on and return results to read the full text of the law. ElasticSearch powers both of these endpoints, as shown in Figure \ref{fig:databaselayout}. This flow is useful for when journalists want to explore a specific term or concept irrespective of its discourse role, or simply familiarize themselves with the corpus.

In \textbf{Flow 2}, users can view aggregate counts of different discourse elements, by type, across the corpora. This helps to summarize the corpora from a functional standpoint, as described in Section~\ref{sct:background}. Users navigate this flow by clicking on one of five buttons to see the counts of each of the five principle discourse spans, then clicking on any of the returned span results to view all laws with this span. MySQL serves both of these endpoints (and provides additional metrics, such as a map in the \texttt{\/about.html} page, not shown here.). 

In both flows, visitors can access our annotation framework, described in Section~\ref{sct:annotationinterface}. From \textbf{Flow 1}, they can click search results to tag a specific paragraph, and from \textbf{Flow 2} they can click to correct an annotated paragraph. Additionally, they can annotate a randomly selected paragraph by clicking ``Help Us Tag.''

\section{Use Cases}
\label{sct:usecase}

We describe two example articles that are currently being explored by users of our system. 

In the first example, journalists hypothesized that the allocation of new liquor licenses might be population-based. To explore this, they used \textbf{Flow 1}; they searched for the term ``alcohol OR liquor OR beverage'' in the search interface and discovered that interface returned 270 laws. They reached out to us and, offline, we analyzed the breakdown in states\footnote{With this project, we introduce a web-app which can primarily help journalists explore our models and generate leads for stories -- we do not intend for it to fully answer all analytical questions. As such, there is a back-and-forth with all the journalists we are working with.}. We found that the states most likely to base liquor licenses off population counts were Tennessee, New York and Illinois. They then asked us to extract all TESTS from these laws. We found that mid-size cities would be the most likely to be impacted by a 5\% or 10\% undercount in population. The journalists identified key cities and sought sources in these areas. This work is ongoing.

In another example, journalists explored \textbf{Flow 2}. They noticed that some TESTS are based on explicit population thresholds (ex. Figure \ref{fig:example_law}) and that some of these thresholds were very narrow. They reached out to us. We compiled several keyword filters and regular expressions extract specific population thresholds. We found that, in Tennessee in particular, over 40\% of all Census-related laws imposed narrow population tests of fewer than 500 people and 10\% imposed tests of fewer than 100 people. This raised questions: what is the purpose of these narrowly targeted laws? Were they trying to target specific counties without mentioning them by name? The journalists are now investigating further by tracking down the authors of these laws.

\section{Related Work}

Although the field of AI-driven legal aids is multifaceted and growing \cite{kauffman2020ai}, free and open-source frameworks remain few \cite{slaw2019, dale2019law, vergottini2011}. Our discourse-driven web application, designed for legal exploratory analysis is one of the few AI-powered, free applications that exist, and the first to open source tools for legal document collection. 

For-profit legal inquiry systems, as mentioned above, are numerous. Bloomberg Law\footnote{\url{https://pro.bloomberglaw.com/}}, Westlaw\footnote{\url{https://www.westlaw.com/}}, LexisNexis\footnote{\url{https://www.lexisnexis.com}} and Wolters Kluwar\footnote{\url{https://www.wolterskluwer.com}} are the four main services for legal research \cite{dale2019law}, which provide subscription-based, Google-style searches. CaseText\footnote{\url{https://casetext.com/}} and Ravel\footnote{\url{https://home.ravellaw.com/}} were two upstart case-text search engines (although both have now been aquired); CaseText offered crowdsourced annotations and Ravel linked cases together to create visual maps of important cases \cite{lee2015new}. We similarly provide a way of collecting user-annotations, and a novel way linking together cases, although ours is linguistically grounded in discourse theory.

Various discourse schemas have been developed to understand law texts, including deontological logic-based schemas \cite{wyner2011rule, zeni2015gaiust}, and subject matter-specific schemas \cite{espejo2019end}. Ours is the first discourse-based approach to take steps towards a big-data approach by setting up a framework for the ingestion of crowdsourced annotations.

Finally, outside of the legal domain, other areas have experienced a growth in academically-oriented systems for human-in-the-loop inquiry. The COVID-19 pandemic has produced a burst in NLP-driven corpora-collection \cite{wang-etal-2020-cord}, demonstrations \cite{sohrab-etal-2020-bennerd, hope-etal-2020-scisight, spangher-etal-2020-enabling} and workshops \cite{nlp-covid19-2020-nlp, nlp-covid19-2020-nlp-covid}. 

Such concerted effort in the NLP domain to expose open resources and build free tools for subject matter experts is an inspiring guide for how researchers can contribute to wider inquiries. We hope such acknowledgement of our ability as NLP researchers to empower others becomes commonplace for other applications as well, forming a common alliance between academics, civil-minded journalists and other researchers and end-users.

\section{Conclusion}

In this work, we have presented three open-source components. (1) A web-app highlighting a novel discourse schema and its application to U.S. Census-related state law. (2) A flexible and modular annotation framework that can be seamlessly embedded into web-apps to allow visitors to contribute and update annotations. (3) A set of 25 web-scrapers to help researchers gather public-domain legal text. Our concrete goal is to facilitate journalistic exploration into laws possibly affected by the 2020 U.S. Census, of which we have multiple ongoing projects already. Our longer-term goal is to collect feedback and data, and improve our database and machine learning systems. We hope that such efforts can continue to push Legaltech into a more open and accessible domain, and make it easier to understand the laws governing our society.

\section{Impact Statement}

There were several possible ethical considerations we encountered during this research which we wish to address. 

\textbf{Dataset Creation}: The creation of our dataset involved scraping numerous websites, including state websites, state-licensed LexisNexis pages and \url{https://www.justia.com}. In the third case, Justia, we did not violate any terms of service. In fact, Justia's \texttt{robots.txt} file\footnote{Found here \url{https://www.justia.com/robots.txt}. Such files govern the site-owners' standards for scraping and crawling.} is the most permissive possible, giving unlimited license to any crawler. It is generally accepted that \texttt{robots.txt} files are implied licenses of access,\footnote{\url{https://stackoverflow.com/questions/999056/ethics-of-robots-txt}} and we did not disregard Justia's file before scraping.

Content derived from the first two categories, state law websites and official, state-licensed websites like LexisNexis are, by law, public domain \cite{wolfe2019, macwright2013}. Web-scraping the public domain is neither illegal nor unethical \cite{mehta2021}. As we did in the body of the paper, we again emphatically criticize attempts by providers to make web-scraping difficult, and we went to lengths to overcome this.

\textbf{Dataset Annotation}: All parties involved in annotating our dataset received valid compensation. We relied entirely on expert researchers to collect our annotations. This included the authors of this paper. All the researchers who provided annotations for us were affiliated with our institution and compensated appropriately by our institution (we leave the determination of ``appropriate'' for our institution to define). 

Although we describe accommodating AMT tasks in the body of the paper, thus far, we have not used any annotations made by Turkers on AMT or or by journalists/researchers using our site. If we do, we will ensure there are no ethical issues by securing university IRB approval or exemption, as deemed fit by the IRB. For the Turkers, we will calculate a payment that equals, on average, \$15 an hour. For the journalists/researchers, we will have exchanged something of value (the use of our web-app) for the annotation.

\textbf{Website Usage}: Our website has significant accessibility limitations for the seeing-impaired and for non-English speakers. We have not addressed them in this current version, but are mindful and actively searching for options to expand accessibility.

There are two ways in which seeing-impaired users might suffer. First, blind users will not be able to read any of the site without external tools, as we have not recorded or built in any native audio-scripts, keyboard shortcuts or voice-activated commands. Secondly, part of our website, \textbf{Flow 2}, introduces users to our discourse schema by introducing them to . The web-app does not attempt to natively include accessibility options for seeing-impaired users.

Our website focuses on U.S.-based laws and contains only English-language text. We do not attempt, in this version, to perform translations. Our plan in the present iteration of this work was to work with U.S.-based journalists studying U.S.-based law. We have not yet undertaken a study to compare how well our discourse schema would apply to non-U.S. law, be it common or civil \cite{dainow1966civil}. However, if this approach proves useful for journalists and researchers, we will certainly seek to undertake this.

\bibliographystyle{acl_natbib}
\bibliography{custom}
\end{document}